\documentclass[
]{ceurart}

\usepackage{listings}
\begin{document}

\copyrightyear{2026}
\copyrightclause{Copyright for this paper by its authors.
  Use permitted under Creative Commons License Attribution 4.0
  International (CC BY 4.0).}

\conference{GeoExT 2026: Fourth International Workshop on Geographic Information Extraction from Texts at ECIR 2026, April 2, 2026, Delft, The Netherlands}

\title{Geolocating News about Extreme Climate Events: A Comparative Analysis of Off-the-Shelf Tools for Toponym Identification in German}


\author[1,2]{Brielen Madureira}[%
email=brielen.madureira@uni-leipzig.de,
]
\cormark[1]
\address[1]{LeipzigLab - Climate Discourse, Leipzig University, Leipzig, Germany}
\address[2]{Helmholtz Centre for Environmental Research - UFZ, Leipzig, Germany}

\author[2]{Mariana Madruga de Brito}[%
email=mariana.brito@ufz.de,
]

\author[1,3]{Andreas Niekler}[%
email=aniekler@informatik.uni-leipzig.de,
]
\address[3]{Computational Humanities, Leipzig University, Leipzig, Germany}


\begin{abstract}
  Determining the geolocation of extreme climate events and disasters in texts is a common problem in climate impact and adaptation research. Named-entity recognition (NER) tools are typically used to identify a pool of toponyms that serve as candidate event locations. In this study, we conduct a comparative analysis of three off-the-shelf NER tools, namely Flair, Spacy and Stanza. We describe and quantify differences between their outputs for German news articles and evaluate them extrinsically based on three methods to determine the country where events took place. We show how their contrasts are propagated into downstream tasks and can yield distinct decisions about a document's geographical focus, which, in turn, can impact conclusions about countries' prominence in German media.
\end{abstract}


\maketitle

\section{Introduction}
\label{sec:intro}

\begin{figure}[h]
	\centering
	\frame{\includegraphics[trim={0cm 13.5cm 0 0cm},clip,width=1\textwidth]{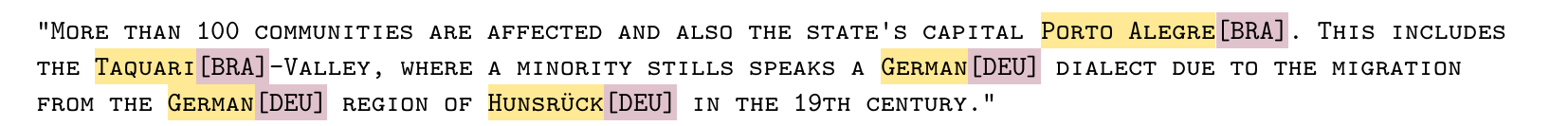}}
	\caption{Excerpt from a German news article (translated by the authors)\protect\footnotemark~reporting on an extreme climate event, annotated with toponym candidates (detected by Stanza's German NER tool) and their country. Highlighted entities must be correctly parsed to determine that Brazil, and not Germany, is the actual event's location.}
	\label{fig:example}
\end{figure}

\footnotetext{Source: Deutsche Welle on May 3, 2024. Available at \url{https://www.dw.com/de/sturzregen-w\%C3\%BCtet-seit-tagen-im-s\%C3\%BCden-brasiliens/a-68988756}.} 

The floods in South Brazil in May, 2024 were so disastrous that they found their way to international news, as shown in Figure \ref{fig:example}. Many human readers have the needed linguistic and cognitive skills to easily realise that the event location in this example is \textit{not} in Germany, despite the referring expressions alluding to this country. But for computational tools this distinction is not so simple: they can be misled by the presence of Germany-related terms detected as toponyms (i.e.~``named entities that label a particular location'' \citep{gritta_pragmatic_2020}). Indeed, \textit{accurate} automation of document geolocation remains an unsolved problem, even for deep learning-based solutions \citep{hu_location_2024}.

Corpora of news articles have been increasingly used to analyse media coverage of climate-related events and extract information about their impacts \citep{otto_fixed_2022,sodoge_automatized_2023,lochner_climate_2024,henrique_lima_alencar_flash_2024,kong_analyzing_2025}. As climate extremes are inherently place-specific, accurate and precise location information is essential for  meaningful spatial analyses. This often involves determining the so-called geographical scope (or geospatial focus) of a document, which is a well-established task in the literature \citep{amitay_web--where_2004,andogah_every_2012,monteiro_survey_2016,melo_automated_2017}. It goes beyond enriching texts with token-based geolocation: after performing toponym identification and resolution, the metadata must be used to infer the actual location of the event of interest \cite{zong_assigning_2005,lee_lost_2019}. Such a task cannot be realistically performed at scale without the aid of automated NLP tools. 

The initial subtask of identifying all toponyms in the text is usually accomplished by named-entity recognition (NER) tools that treat locations as a specific entity. When this task is of secondary importance in a project, researchers often rely on off-the-shelf NER tools, taking their correctness at face value. However, such tools are not error-free and their inaccuracies can (possibly unnoticeably) are propagated to downstream decisions and results.

To investigate that, we conduct a comparative analysis of the toponyms identified by three popular off-the-shelf NER tools in a corpus of German news articles about various disasters and extreme climate events (e.g.~floods, heat waves, fires and droughts). We quantify variations in their outputs and compare three prediction methods to show how these differences influence the determination of documents' geographical focus and, consequently, higher-level conclusions about the prominence of countries in German media.

\section{Related work}
\label{sec:lit}
NER typically subsumes toponym identification as it usually includes specific labels for locations. There is vast literature about NER tools and datasets for German texts \citep[\textit{inter alia}]{benikova-etal-2014-nosta,riedl-pado-2018-named,labusch-etal-2019-bert,leitner-etal-2020-dataset}. However, the performance of off-the-shelf natural language processing (NLP) tools is not always optimal, particularly for languages other than English. Systematic evaluation is therefore essential to establish a robust and context-sensitive plan of action for each use case and domain. Previous studies, such as \cite{ortmann_evaluating_2019,scheible-etal-2011-evaluating,laarmann-quante-etal-2022-evaluating}, assessed the performance of available tools for various NLP tasks in German. Yet, to our knowledge, recent comparative studies focusing on NER tools for German are missing from the literature. 

A similar gap exists in the context of geotagging. While toponym identification has been widely evaluated, most empirical studies rely on English-language datasets \citep{gritta_whats_2018,wang_enhancing_2019,liu_geoparsing_2022,hu_location_2024}. Although existing tools can correctly identify many toponyms, they also incur false positives and false negatives. In their practical guide to geoparsing evaluation, \cite{gritta_pragmatic_2020} argue that ``off-the-shelf NER taggers are inadequate for location extraction''. Common sources of error include the lack of distinction between pragmatic types of toponyms, limited handling of metonyms, entity boundary errors and case sensitivity issues \citep{gritta_whats_2018,gritta_pragmatic_2020}. Moreover, German climate-related news poses its own challenges \citep{wohlgemuth_geo-parser_2025}.  

Several studies use a gold standard as a reference for evaluating toponym identification. For instance, \cite{won_ensemble_2018} compared the performance of various NER tools for historical corpora in English to manually annotated toponyms and, more recently, \cite{kriesch_geolocated_2025} used a large language model to annotate German news, using the resulting dataset to train and evaluate a custom NER model. However, in real-world settings, annotated samples of comparable data (e.g.~of same genre, domain and region) are not always available. 

Our work builds on the existing literature but addresses a particular situation in which off-the-shelf NER tools must be selected and used without an available gold standard for evaluation. To assess their performance under these constraints, we adopt an extrinsic evaluation approach that measures how reliably the geotagged toponyms of each tool can be used for document-level geolocation and frequency-based ranking of countries in a corpus of German news.

\section{Methods}
\label{sec:method}
\paragraph{Task} Our task is a special case of geographical focus identification, similar to \cite{kong_analyzing_2025}: given a collection of news articles reporting on or mentioning disasters and extreme climate events, our goal is to determine the country (or countries) in which the event(s) of interest occurred.  This task is a step towards data-driven analyses of the coverage of worldwide climate events in German newspapers.

\paragraph{Procedure} We perform this task in a 4-step pipeline as described below. Here, we do \textit{not} aim to propose a ground-breaking method for document-level geolocation. Instead, the main contribution lies in the evaluation: comparing NER tools and examining whether they yield diverging downstream conclusions. For that, steps 1-4 rely on basic heuristics inspired by existing literature but without sophisticated optimisation. The deliberate use of simpler methods ought to avoid obscuring the effects of NER decisions behind more complex modelling choices.

\begin{enumerate}
	\item \textbf{Identification} Each document in the corpus is enriched with an annotation layer for each NER tool indicating tokens they identify as candidate toponyms (i.e.~labeled as \texttt{location}).
	
	\item \textbf{Querying} Each unique toponym type identified over the whole corpus is used to build a query for a database of geographical coordinates. The top $n$ matches are retrieved with their metadata, which includes (when available) latitude, longitude and country.
	
	\item \textbf{Resolution} Each toponym type in each document is mapped to its likely geographical coordinates. When multiple coordinates were retrieved for a toponym type, the ambiguity is resolved by taking the coordinates that are closest to the polygon formed by the unambiguous toponyms in the document. This approach is based on the ``spatial minimality'' heuristic used by \citep{leidner_grounding_2003} and the clustering approach by \citep{habib-2013}. If all toponyms are ambiguous, the most probable match (i.e.~the one ranked first by the database's search engine) is selected for each toponym instead. 
	
	\item \textbf{Prediction} For each tool and geographical database pair, we apply three prediction methods that get a document's toponyms and coordinates as input and map them to countries.
	
\end{enumerate}

\paragraph{Evaluation} The NER tools' performance is first assessed \textit{bilaterally}, by intercomparison of their outputs, and then \textit{extrinsically}, by their effect on subsequent tasks of document geolocation and frequency ranking, using human annotation as a gold standard for the documents' geographical focus. We also examine characteristics of the outputs that may explain the observed behaviour.

\section{Experiments}
\label{sec:experiments}
This section presents an overview of the design decisions for our experiments regarding five main dimensions: data, NER tool, geographical database, country prediction method and evaluation metric. Further details can be found in the Appendix. Our implementation is available for documentation purposes at \url{https://codeberg.org/briemadu/german-ner-geoparsing}.

\paragraph{Data} Our dataset is derived from an ongoing research project about climate discourse in the German media. It is comprised of 983 documents from German newspapers published between 2000 and 2024 retrieved from the \texttt{wiso-net}\footnote{\url{https://www.wiso-net.de/}} news aggregator database using a selection of keywords related to seven hazards (heat wave, wildfire, flood, storm, drought, cold wave and landslide). The texts are in German and were annotated by humans with the type of event they discuss and the country (or countries) where the event of interest happened. The number of characters per text ranges from 178 to 11,512 with, on average, 2,468.

\paragraph{Models} The analysis assesses three popular off-the-shelf tools that perform NER in German: Flair \citep{akbik-etal-2019-flair}, using model \texttt{de-ner-large},\footnote{ \url{https://flairnlp.github.io/docs/tutorial-basics/tagging-entities}} Spacy \citep{honnibal2020spacy}, using model \texttt{de\_core\_news\_lg},\footnote{\url{https://spacy.io/models/de\#de_core_news_lg}} and Stanza \citep{qi-etal-2020-stanza}, using model \texttt{de}.\footnote{ \url{https://stanfordnlp.github.io/stanza/ner_models.html}} All tools were trained with four labels (for person, location, organisation and other). In this study, only the label \texttt{LOC} for location is relevant. Using LLMs is also a possibility, but not the focus of this study, as they are not NER tools \textit{by design}.

\paragraph{Geographical database} The latitude and longitude coordinates with their respective country for each toponym type are retrieved from two geographical databases: Geonames and Nominatim.\footnote{\url{https://www.geonames.org/} and \url{https://nominatim.org/}, respectively.}

\paragraph{Prediction methods} Three methods are used to determine the geographical focus (here, the country where the discussed event occurred) of a document based on its identified toponyms:

\begin{enumerate}[i]
	\item \textbf{Majority voting}: the most common country (or countries, if there is a tie) among the pool of toponyms' countries in a document is (are) selected as its geographical focus. This is inspired by voting methods used in other tasks \citep{hu_location_2024} and the ``popularity approach'' by \citep{habib-2013}.
	
	\item \textbf{Closest to centroid}: the centroid of the concave hull of the polygon formed by the document's toponyms, after removing outliers, is computed. The toponym closest to its centroid is selected as the document's geographical focus. This is similar to the approaches by \cite{goos_disambiguating_2001,pasley_geo-tagging_2007,radke_geotagging_2018}.
	
	\item \textbf{Keyword proximity}: the list of a document's toponyms is reduced to only those that co-occur in a sentence with a hazard keyword (the same used to query documents) before applying majority voting. If no sentences contain both a keyword and a toponym, we resort to the closest toponyms (in number of tokens) that comes preferably before each keyword (or after, in case none precedes it).
\end{enumerate} 

\paragraph{Evaluation metrics} The intersection over union (IoU) metric (or Jaccard Index) is used to measure how much the set of identified toponyms per document coincide between pairs of tools, considering only exact matches (i.e.~same initial and last character). Predicted countries are assessed in relation to the gold standard using the percentage of exact matches (when the exact same country or countries were predicted) and the percentage of partial matches (when at least one correct country was predicted). Finally, Kendall $\tau$ and Spearman’s rank correlation coefficients \citep{spearman_proof_1904,kendall1938new} are used to compare the resulting rankings of country frequencies in the corpus.

\section{Analysis}
\label{sec:analysis}
The analysis is divided into four parts. We first compare the sets of toponyms identified by the three NER tools, considering a realistic situation in which gold standard toponyms are not available. Then, we examine how effective each tool's outputs are for two higher level tasks using our annotation: determining the document's geographical focus and assessing the ranked frequency of each country in the corpus, which could inform conclusions about countries' prominence in the media. We finish by discussing properties of the NER outputs that may explain some of the observed variations.

\subsection{Toponym identification: bilateral comparisons} 

	In total, 9,231 toponym types were identified in the corpus, among which 5,836 (63.22\%) had matches in Nominatim and 3,613 (39.14\%) were found in Geonames. Table \ref{table:stats} shows the number of toponyms and of toponym types identified by each tool over the whole corpus, as well as the percentage of types that returned at least one valid match in the geographical databases. Although Spacy identifies most toponyms and most unique types, it has the worst coverage in both databases. Flair, on the other hand, identified fewer toponym types but with a higher percentage of valid ones.
	
	\begin{table}[h]
		\centering
		\begin{tabular}{rcccc}
			\toprule
			& $n$ toponyms & $n$ types & \% in Geonames & \% in Nominatim \\
			\cmidrule{1-5}
			Flair & 11,816 & 4,639 & 63.16  & 81.20 \\
			Stanza & 14,230 & 5,330 & 54.11 & 77.97 \\
			Spacy & 14,367 & 6,700 & 42.75 & 63.58 \\
			\bottomrule
		\end{tabular}
		\caption{Overview of the number of toponyms identified by each tool and \% of matches in geographical databases.}
		\label{table:stats}
	\end{table}
	
\begin{figure}[h]
		\centering
		\includegraphics[width=0.45\textwidth]{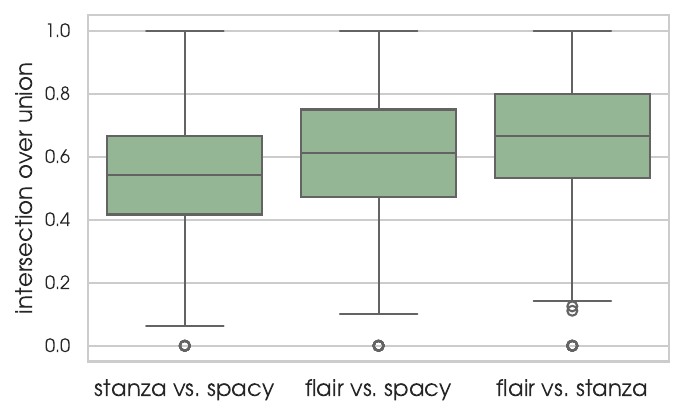}
		\caption{Box plots of toponyms' intersection-over-union values per document for each pair of tools.}
		\label{fig:iou}
	\end{figure}

	Figure \ref{fig:iou} shows the distribution of intersection-over-union values computed for the sets of toponyms per document for each pair of models. In our dataset, Spacy and Stanza tended to have more disagreements, whereas Stanza and Flair tended to agree comparatively more. While the three medians are similar, the important insight is that most values fall between 0.4 and 0.8, indicating that, on average, there is considerable variation in the set of toponyms identified by different tools for the same document. 
	
\subsection{Higher-level task: country prediction} 

	Moving on to the main task of determining the geographical focus of each document, we first compare NER tools bilaterally, without the gold standard. Table \ref{table:bilateral} presents the percentage of exact agreement between the prediction of countries based on the toponyms identified by each tool. Again, Spacy and Stanza incur the most disagreements, whereas Stanza and Flair agree the most in all but one experiment. 
	
	\begin{table}[h]
		\centering
		\begin{tabular}{rcccccc}
			\toprule
			& \multicolumn{3}{c}{Geonames} & \multicolumn{3}{c}{Nominatim} \\
			\cmidrule(lr){2-4}  \cmidrule(lr){5-7}
			& centroid & keyword & majority & centroid & keyword & majority \\
			\cmidrule(lr){2-4}  \cmidrule(lr){5-7}
			
			Stanza vs.~Flair & 80.47 & 87.69 & 89.01 & 72.84 & 83.01 & 82.40 \\ 
			Spacy vs.~Flair & 76.50 & 82.60 & 82.91 & 70.50 & 82.50 & 83.01 \\
			Spacy vs.~Stanza & 72.74 & 80.98 & 81.49 & 65.41 & 77.11 & 79.86 \\
			
			\cline{1-7}
			\bottomrule
		\end{tabular}
		\caption{Percentage of exact agreement among documents' predicted countries for each model pair and method.}
		\label{table:bilateral}
	\end{table}
	
	These results show that the variations in the toponyms identified by each NER tool caused at least 10\% (but in some cases more than 20\%) of the country predictions to be different, a consistent finding across prediction methods and geographical databases.
	
	To assess the actual performance in this task, the next step is comparing them with the gold-standard. Table \ref{table:gold} presents the percentage of exact and partial matches for country predictions based on each tool's outputs and method. We observe that, for our data, predictions using toponyms near hazard-related keywords led to the best overall performance in most cases, followed by majority voting. Additionally, Nominatim was more effective than Geonames in general. However, we reiterate that our aim here is not yet to find the best prediction method for solving the problem. The different methods are a means to assess the NER \textit{tools}, and to what extent their distinction is consistent across varied experiments. 
	
	For all prediction methods and both geographical databases, Flair had the highest results, followed (in most cases) by Stanza. The main finding here is that, \textit{ceteris paribus}, there can be a difference of up to 5 p.p.~in the accuracy of the predictions which is only due to the mere choice of NER tool. Therefore, if only one tool had been used directly without careful inspection and comparison, the performance of the predictions methods in this downstream task could have suffered an unwarranted negative impact.
	
	\begin{table}[h!]
		\centering
		\begin{tabular}{ll|cc|cc}
			\toprule
			\multicolumn{2}{c}{}  & \multicolumn{2}{c}{Geonames} & \multicolumn{2}{c}{Nominatim} \\
			\cmidrule(lr){3-4} \cmidrule(lr){5-6}
			\multicolumn{2}{c}{}  & exact & \multicolumn{1}{c}{overlap} & exact & overlap \\
			\cmidrule{1-6}
			
			\multirow[t]{3}{*}{keyword} & Flair & 62.67 & 75.38 & 70.60 & 83.01 \\
			& Stanza & 59.61 & 74.06 & 65.11 & 81.18 \\
			& Spacy & 57.07 & 70.80 & 65.82 & 79.86 \\
			\cmidrule{1-6}
	
			\multirow[t]{3}{*}{majority} & Flair & 60.73 & 73.14 & 65.01 & 78.33 \\
			& Stanza & 58.70 & 72.23 & 61.95 & 74.47 \\
			& Spacy & 57.58 & 71.31 & 62.77 & 77.62 \\
			\cmidrule{1-6}
	
			\multirow[t]{3}{*}{centroid} & Flair & - & 63.68 & - & 66.84 \\
			& Stanza & - & 62.46 & - & 61.95 \\
			& Spacy & - & 60.22 & - & 62.26 \\
	
			\cline{1-6}
			\bottomrule
		\end{tabular}
		\caption{Percentage of exact and overlapping matches of predictions based on each tool in relation to the gold standard. Exact match is not presented for the centroid method because it does not predict more than one country.}
		\label{table:gold}
	\end{table}

\subsection{Higher-level task: ranking by prominence} 

	In our pipeline, NER is used to guide predictions which, in turn, are supposed to be used for even higher-level conclusions. Assume we are interested in ranking countries by media prominence and consider the number of documents about a country in the corpus as a proxy for this construct. We thus order countries by how many documents were mapped to them in the gold standard and by each tool. 
	
	This part of the evaluation is conducted using results from the keyword method with Nominatim, which achieved the best metric values in determining the geographical focus so far. Table \ref{table:corr} shows the Spearman rank correlation and the Kendall's $\tau$ for the rankings when considering only the countries that have at least 10 documents in the gold standard. While Spacy and Stanza's exhibited very similar correlations with the reference, Flair reached a higher   correlation.\footnote{This difference reduces if lower rank positions are used, but the metrics are less reliable due to the small number of instances.} 
	
	\begin{table}[h!]
		\centering
		\begin{tabular}{lrrr  lrrr}
			\toprule
			& Spacy & Stanza & Flair & & Spacy & Stanza & Flair \\
			\cmidrule(lr){2-4} \cmidrule(lr){6-8}
			Spearman r & 0.753 & 0.758 & 0.833 & \hspace*{1cm} Kendall $\tau$ & 0.648 & 0.641 & 0.714 \\
			\bottomrule
		\end{tabular}
		
		\caption{Correlation between the predicted and observed number of documents per country (with at least 10 instances).}
		\label{table:corr}
	\end{table}
	
\begin{figure}[h]
		\centering
		\includegraphics[width=0.5\textwidth]{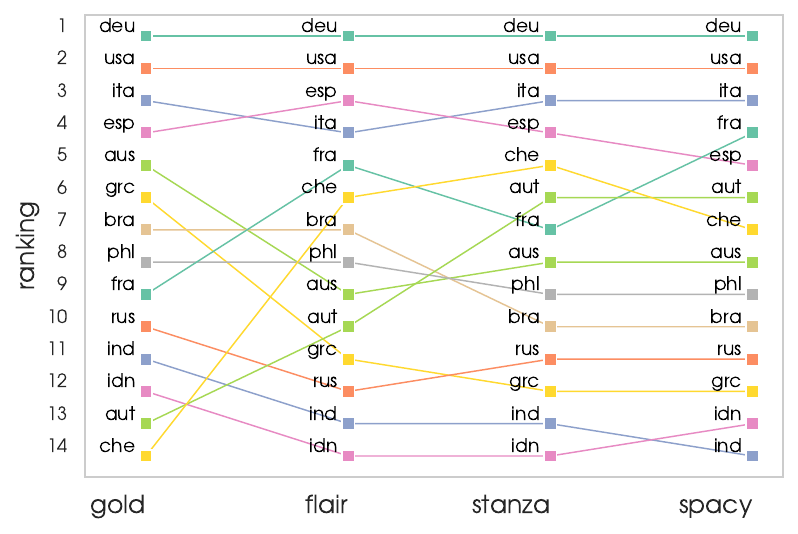}
		\caption{The top 14 countries with the highest frequency in the gold standard ranked using each NER tool.}
		\label{fig:ranking}
	\end{figure}

	Figure \ref{fig:ranking} shows the top 14 countries ranked by the number of documents that have them as geographical focus, for NER tools and the gold standard. All tools ranked the same set of 14 countries at the top, and the two most frequent countries were also correctly determined. This is evidence that good overall agreement in identifying the most prominent countries could be reached despite the contrasts in the tools' outputs. Still, the country order varies. For instance, Switzerland (\texttt{che}) ranks much higher across all tools than it does in reality in the corpus. Stanza and Spacy lead to similar conclusions for most countries: 9 of them appear in identical positions and 3 are in adjacent positions. The main discrepancy between these two tools is that Spacy ranks France (\texttt{fra}) 3 positions higher than Stanza. Flair and Stanza, on the other hand, coincided only in 4 positions. The Kendall correlation for this portion of the ranking was 0.89 between Stanza and Spacy, 0.78 between Flair and Stanza and 0.75 between Flair and Spacy. Stanza and Spacy agreed more with each other, but Flair was superior in relation to the gold standard (correlation of 0.53, as opposed to 0.47 and 0.45 of Spacy and Stanza, respectively).

\subsection{Error analysis}

To conclude the analysis, we investigate which properties of each NER tool's outputs may have contributed to the different predictions.
First, we note that average number of toponyms per document was higher for Spacy (mean $14.61$, sd $12.74$) and Stanza (mean $14.47$, sd $12.25$) than for Flair (mean $12.02$, sd $10.63$). Part of these extra toponyms may have contributed to divergent predictions, as two extra tokens can already change majority voting or a polygon's centroid, for example.

Looking closely, among the 9,231 toponym types identified in the corpus, 683 (7.40\%) were unique to Flair,  1,276 (13.82\%) were unique to Stanza and 2,847 (30.84\%) were unique to Spacy.
All of Flair's unique toponyms occur no more than 5 times in the corpus and may have had a lower impact on our outcomes (although this can also become a problem in larger corpora).
Among the most common types identified only by Spacy, we see detached generic terms like \textit{Stadt} (city), \textit{Altstadt} (old town), \textit{Innenstadt} (downtown), \textit{Flughafen} (airport), \textit{Provinzen} (provinces), \textit{Gemeinde} (community) and \textit{Rathaus} (town hall). Among the top unique toponyms identified by Stanza, apart from the word \textit{Sonne} (sun) that appears very often, we find mostly adjectival and demonymic forms of country and city names, classified as \textit{associative toponyms} in the taxonomy proposed by \cite{gritta_pragmatic_2020}.  
Although these forms often do not label a specific place, some of them correspond to location names in different countries, which introduces a source of noise into the methods.
Associative toponyms may even map to countries not associated with the toponym itself (for instance, \textit{italienischen}, the declined form of the adjective Italian, triggered place names like embassies in six countries--and none of them was even in Italy). It is in fact debatable to what extent associative toponyms should be included (see \cite{gritta_pragmatic_2020}). 

There is mutual influence between geographical databases and NER outputs. On the one hand, the limited coverage of toponyms in geographical databases (Table \ref{table:stats}) may be a sign that they help safeguard the subsequent use of tools against their invalid toponyms. But the spurious multiple matches for some underspecified place names in different countries require a more fine-grained treatment. The labelling scheme also contributes to the drawbacks: in some other languages, NER tools are trained with more location labels that enable a distinction between geopolitical entities, nationalities/affiliations, facilities and locations. German is impaired by a single label that encompasses various concepts. A possibility is to filter identified toponyms by their part-of-speech tag and syntactic role in their sentence.

\section{Conclusion}
\label{sec:conclusion}
We conducted a comparative analysis of three off-the-shelf NER tools for German applied to news articles about disasters and extreme climate events. In a setting where toponym annotation was not available, we first assessed the tools' performance by how closely they aligned with one another and then by their extrinsic performance on higher-level decisions. Flair achieved the best results in our data, in line with its superior F1 score reported by the provider. Stanza and Spacy were comparable, with some advantage to one or to the other depending on the experiment. Nevertheless, we avoid making general claims about which is inherently superior and thus did not statistically test any hypothesis in that regard. \cite{won_ensemble_2018} has actually shown that using a majority voting ensemble of NER tools can improve toponym identification and that tools' performance may be corpus dependent.

What our experiments corroborate is that tool selection can discernibly influence downstream results. Should we blindly decide to use a single tool without comparative experiments, more than 10\% of the predictions could differ, accuracy could be 5p.p.~lower and the resulting ranking could be less correlated to the real observations, depending on the method. These findings provide empirical evidence that set-up decisions which may appear minor at first glance can still affect key conclusions. This underscores the need for careful evaluation and error analysis of off-the-shelf NER tools, even those widely accepted in the community, as their performance can vary in non-negligible ways across specific use cases. 

\begin{acknowledgments}
	
	We thank Maike Reichel and Julius Hehenkamp for their valuable help in annotating the data.
		
\end{acknowledgments}

\section*{Declaration on Generative AI}
  The authors have not directly employed any Generative AI tools to write this paper. One of the authors uses Grammarly for grammar and spelling checks.

\appendix

\section*{Appendix: Implementation Details}
\label{sec:appendix}
\paragraph{Data} Our corpus of news documents was queried using keywords related to seven types of climate extremes, similar to \citep{li-etal-2024-using-llms,flood-tms,Carvalho2025}. Two student assistants annotated a sample of 3,150 documents, 450 for each hazard and, among them, 18 for each year. In this study, we used only the subset of 1,008 documents identified as relevant, i.e.~those that do discuss an extreme climate event. We excluded 25 documents that either do not contain any hazard-related keywords in their main text (because title and subtitle were not considered here) or have no identifiable countries (e.g.~refer to Europe as a whole). 

\paragraph{NER} We extracted all sequences of tokens corresponding to an entity with the label \texttt{LOC}, together with its start and end characters. 
For Stanza, we loaded only the \texttt{tokenize},  \texttt{mwt} and \texttt{ner} components.

\paragraph{Geographical databases} Nominatim was queried via the Python library \href{https://github.com/geopy/geopy}{\texttt{geopy.geocoders}} and Geonames via the library \href{https://github.com/DenisCarriere/geocoder}{\texttt{geocoder}}. For each query, we cached a maximum of 20 matches with the needed metadata: latitude, longitude, country code, type and rank position in the search results. For Geonames, we additionally passed the parameters \texttt{name\_equals} with the toponym string and set \texttt{fuzzy}$=1$. This resulted in 34,803 instances for Nominatim and 30,836 for Geonames. Toponym matches without a known country were filtered out, so that 90.0\% and 81.68\% were retained respectively. For duplicates with the same name and country, we kept only the first match (in our use case, only the country is relevant, without the need for a precise geolocation within it). The final list had 11,300 toponym instances (32.47\% of the initial results) for Nominatim and 10,392 (33.70\%) for Geonames.

\paragraph{Prediction} For majority voting, we counted and ranked the number of toponyms for each country. If there was a tie, we returned all countries that shared the top position. This method accounted for all repetitions of toponym types. In the keyword method, we first selected only toponyms that co-occur in the same sentence with one of the keywords. If there was none, we selected the toponyms closest to each keyword in number of tokens, preferably before, but after if there was none. Repeated keywords and types were also considered. For the centroid method, we first turned each coordinate into a Point object and sorted them in counter clockwise order with respect to their mean. Points were used to construct a Polygon object (we resorted back to a Line object when only two points were found or to a Point object when only one was found). Then, if there was more than one point, we computed the center and excluded all outlier points whose absolute z-score of the distance to the centroid is greater than 1. A new polygon was created with the remaining points and its centroid was computed. Finally, we selected the toponym of the point closest to the centroid. In this method, repeated toponyms were automatically aggregated into a single point in the polygon.

\paragraph{Limitations} We did not account for overlapping sequences of tokens when comparing NER outputs, which could increase their agreement but would complicate merging adjacent tokens and partially overlapping sequences of characters. Furthermore, the queries to the geographical databases used only the identified tokens, in isolation, without contextual disambiguation or toponym declination. Ideally, textual context should be taken into account to resolve ambiguity. As toponyms were not lemmatised here, their German declination can also have caused mismatches in the geographical databases.

\bibliography{sample-ceur}

\end{document}